\documentclass[runningheads]{llncs}
\usepackage[T1]{fontenc}
\usepackage{graphicx}
\usepackage{hyperref}
\hypersetup{
    colorlinks=true,
    citecolor=blue,
    linkcolor=blue,
    filecolor=magenta,      
    urlcolor=magenta,
    pdftitle={BiblioPage: A Dataset of Scanned Title Pages for
Bibliographic Metadata Extraction}
}

\usepackage[misc]{ifsym}
\usepackage{caption}
\usepackage{subcaption}
\usepackage{multicol}

\usepackage{lipsum}
\usepackage{blindtext,letltxmacro,xcolor,xparse,colortbl}

\LetLtxMacro{\blindtextblindtext}{\blindtext}
\LetLtxMacro{\blindtextBlindtext}{\Blindtext}

\RenewDocumentCommand{\blindtext}{O{\value{blindtext}}}{%
  \begingroup\color{gray}\blindtextblindtext[#1]\endgroup
}
\RenewDocumentCommand{\Blindtext}{O{\value{blindtext}}O{\value{Blindtext}}}{%
  \begingroup\color{gray}\blindtextBlindtext[#1][#2]\endgroup
}

\definecolor{Gray}{gray}{0.9}
\definecolor{LightCyan}{rgb}{0.88,1,1}

\newcolumntype{g}{>{\columncolor{Gray}}c}
\newcolumntype{b}{>{\columncolor{white}}c}

\begin{document}
\title{BiblioPage: A Dataset of Scanned Title Pages for Bibliographic Metadata Extraction}
\titlerunning{BiblioPage: A Dataset for Bibliographic Metadata Extraction}
%
\author{Jan Kohút (\Letter) \orcidID{0000-0003-0774-8903} \and
Martin Dočekal\orcidID{0000-0002-0580-9357} \and
Michal~Hradiš\orcidID{0000-0002-6364-129X} \and
Marek Vaško\orcidID{0000-0003-1404-4154}}
\authorrunning{J. Kohút et al.}
%

\institute{Faculty of Information Technology, Brno University of Technology, Brno, Czech~Republic \\
\email{ikohut},
\email{idocekal},
\email{ihradis},
\email{ivasko@fit.vutbr.cz}}

\maketitle              
\begin{abstract}

Manual digitization of bibliographic metadata is time consuming and labor intensive, especially for historical and real-world archives with highly variable formatting across documents. 
Despite advances in machine learning, the absence of dedicated datasets for metadata extraction hinders automation.
To address this gap, we introduce BiblioPage, a dataset of scanned title pages annotated with structured bibliographic metadata. The dataset consists of approximately 2,000 monograph title pages collected from 14 Czech libraries, spanning a wide range of publication periods, typographic styles, and layout structures. Each title page is annotated with 16 bibliographic attributes, including title, contributors, and publication metadata, along with precise positional information in the form of bounding boxes.
To extract structured information from this dataset, we evaluated object
detection models such as YOLO and DETR combined with transformer-based OCR, achieving a maximum mAP of 52 and an F1 score of 59.
Additionally, we assess the performance of various visual large language
models, including LlamA 3.2-Vision and GPT-4o, with the best model
reaching an F1 score of 67.
BiblioPage serves as a real-world benchmark for bibliographic metadata extraction, contributing to document understanding, document question answering, and document information extraction.
Dataset and evaluation scripts are availible at: \url{https://github.com/DCGM/biblio-dataset}

\end{abstract}
\section{Introduction}

One of the main tasks in document digitization, including books, magazines, and newspapers, is the extraction of bibliographic metadata. 
This process is typically performed manually, as bibliographic entities are recorded inconsistently across different documents, publishers, regions, and time periods, making automation challenging.
The complexity of this task is further compounded by the contextual dependencies of many bibliographic elements. 
For instance, correctly recording metadata may require familiarity with an entire periodical when processing a single issue or understanding a multi-volume set when digitizing one of its parts. Additionally, another major challenge lies in accurately locating bibliographic information across the pages of a document.

With advances in machine learning, it is reasonable to assume that the extraction of bibliographic metadata could be at least partially automated. 
However, there is a lack of diverse datasets in this field. 
Existing datasets typically cover a limited number of bibliographic entities and focus on relatively homogeneous documents—primarily recent academic publications—rather than a broader range of real-world books, newspapers, and other sources~\cite{bhardwaj2017deepbibx,rizvi2020deepbird}.

We introduce BiblioPage, a dataset of scanned title pages annotated with structured bibliographic metadata. 
It includes title pages from standalone works, primarily books, spanning various time periods and archival sources across the Czech Republic. 
Each page is annotated with 16 bibliographic attributes, such as title, author, publication date, and contributor roles, reflecting diverse typographic styles, historical printing variations, and layout structures. 
The dataset focuses exclusively on title pages, without addressing the localization of bibliographic metadata across multiple pages or the use of broader contextual information from related documents.

Direct utilization of the BiblioPage dataset can support the development of tools for automated bibliographic metadata extraction in archives and libraries. It can also serve as a benchmark for document question answering and document information extraction, providing new bibliographic entities.

\section{Related Work}
Automatic extraction of bibliographic information has been studied for decades, initially focusing on processing text transcriptions. Early methods applied Hidden Markov Models (HMMs) to handle transcription errors~\cite{takasu2003extraction} or extract structured metadata from web text~\cite{geng2004autobib}. Later, Conditional Random Fields (CRFs) improved extraction from unstructured text~\cite{matsuoka2016crf}. With the advent of layout-aware approaches, Support Vector Machines demonstrated improved performance by incorporating spatial features~\cite{granitzer2012comparison}.

Deep learning has since become the dominant approach, particularly for extracting references directly from document images~\cite{bhardwaj2017deepbibx,rizvi2020deepbird}. These methods typically transcribe and segment references before further processing, often using CRF-based tools like ParsCit~\cite{councill2008parscit}. A related but distinct effort leveraged traditional image processing techniques to extract bibliographic information from book covers~\cite{yang1999cover}, though no corresponding dataset was published.

Technically, modern approaches integrate visual, spatial, and textual information through neural models. Convolutional networks were first used for layout detection~\cite{8237584,liu2016ssd,ren2015faster,khanam2024yolov11}, but transformers have since emerged as state-of-the-art. Models such as DLAFormer~\cite{wang2024dlaformer} combine object detection with text regions for improved relation prediction, while hierarchical approaches build structured representations for retrieval~\cite{wang2024detect}. Transformer-based document models further enhance extraction through self-supervised pretraining. LayoutLM and its successors~\cite{xu2020layoutlm,xu2020layoutlmv2,huang2022layoutlmv3} integrate spatial embeddings with text, DiT~\cite{li2022dit} learns from large-scale unlabeled images, and DocFormer~\cite{appalaraju2021docformer} uses multimodal attention layers. Language-independent methods like LiLT~\cite{wang2022lilt} extend these techniques across different languages.

To support these advances, various datasets facilitate document-level information extraction, though not specifically for bibliographic data. FUNSD~\cite{funsd} focuses on key-value pair extraction in handwritten and printed forms, while XFUND~\cite{xfund} extends this to multilingual settings, adding challenges in script variation. SROIE~\cite{sroie} and CORD~\cite{cord} address structured data extraction from receipts, emphasizing robustness to layout variability and OCR noise. Kleister-NDA~\cite{stanislawek2021kleister} shifts the focus to long-form legal documents, highlighting entity linking and retrieval across pages. While these datasets target different document types, they share methodological similarities with bibliographic metadata extraction.

Another group of relevant datasets originates from the Visual Question Answering (VQA) task~\cite{antol2015vqa}.
While the task originally focused on natural images, community interest turned to processing images with text~\cite{singh2019reading}, eventually resulting in establishment of Document VQA~\cite{Mathew2020DocVQAAD} as a separate task.
While the DocVQA dataset does not focus on bibliographic information either, one can see that the task is in principle closely related.

\section{Dataset}

\begin{figure}[t]
    \centering
    \begin{subfigure}{0.22\linewidth}
        \centering
        \includegraphics[width=\linewidth]{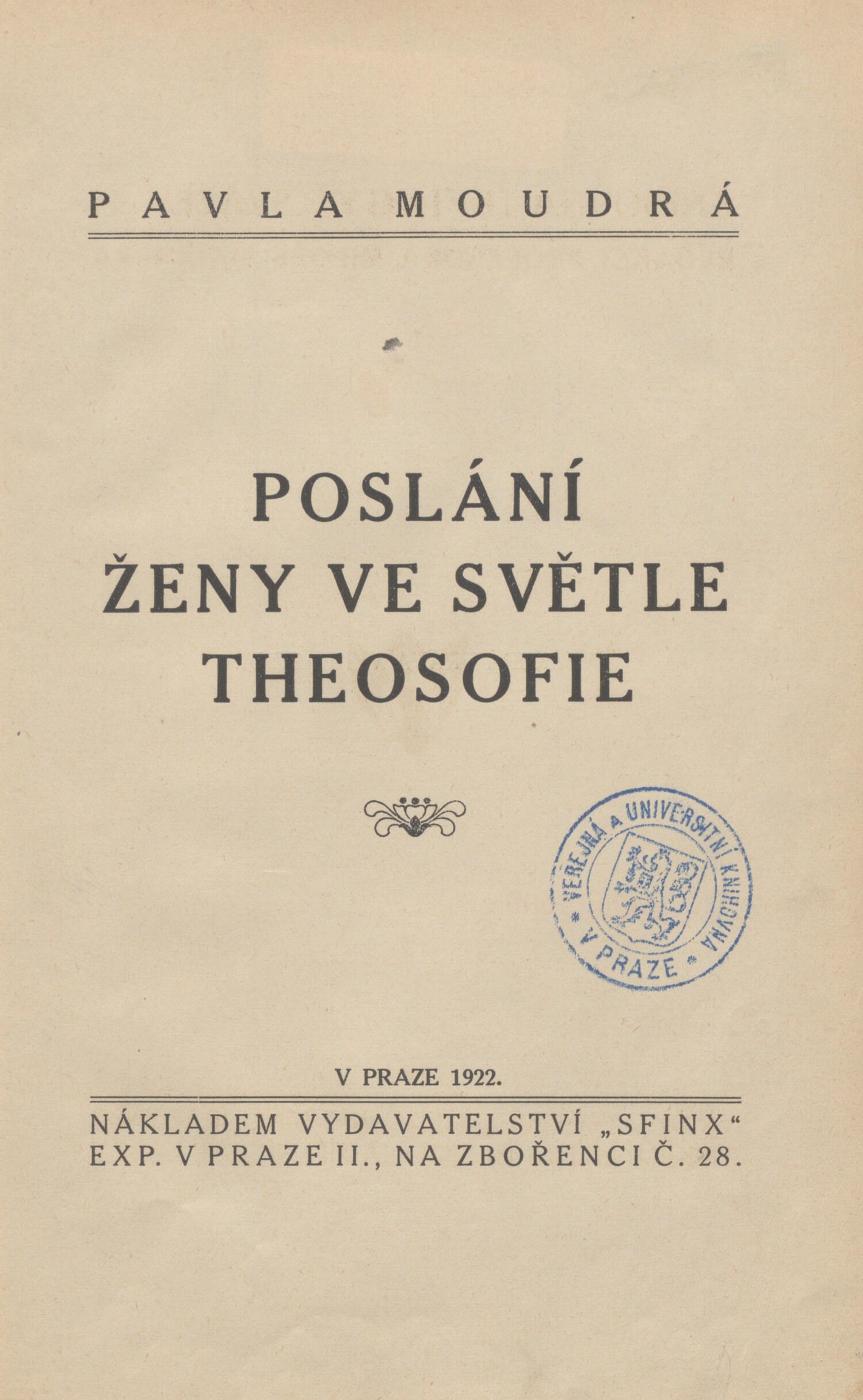}
    \end{subfigure}
    \begin{subfigure}{0.217\linewidth}
        \centering
        \includegraphics[width=\linewidth]{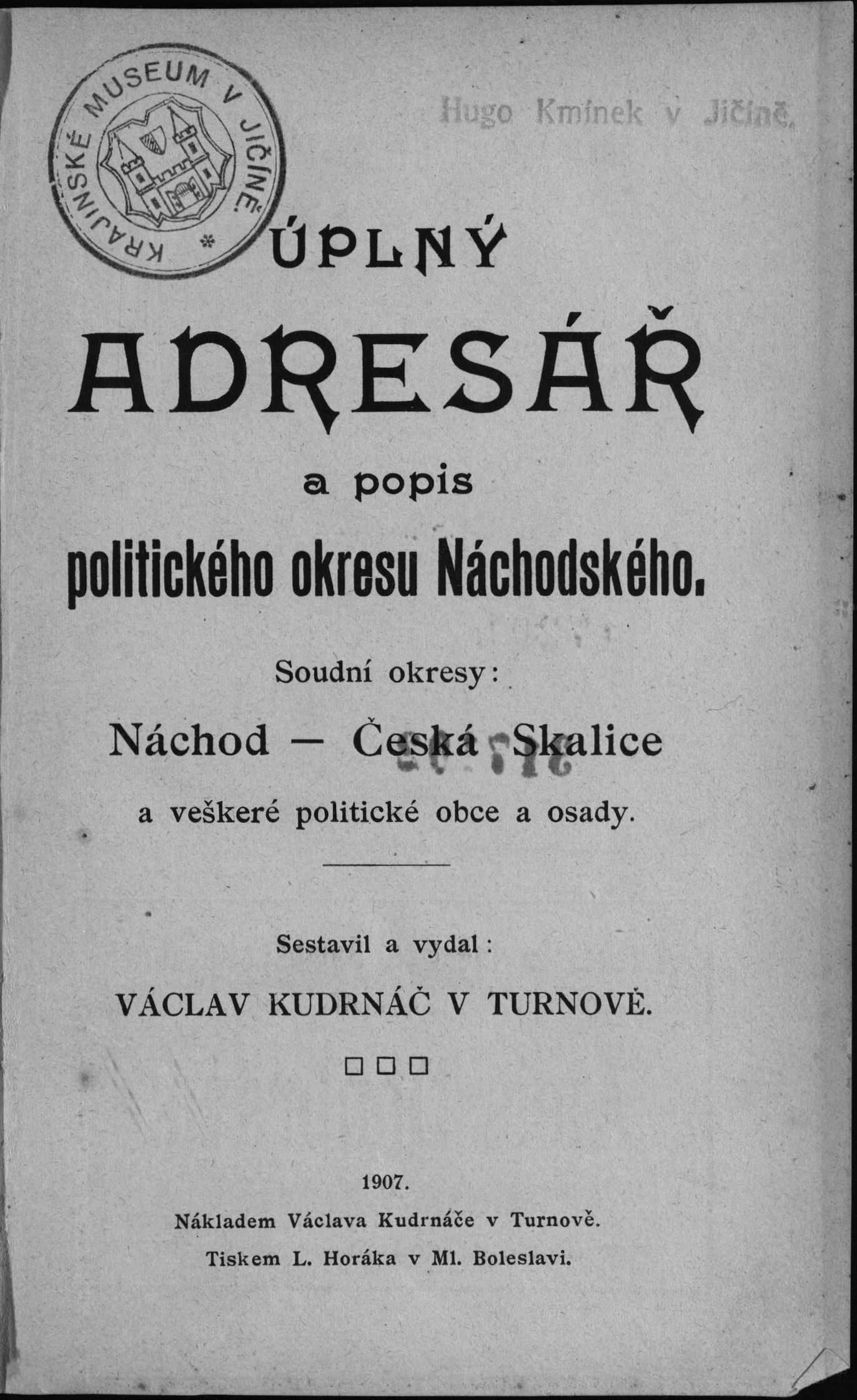}
    \end{subfigure}
    \begin{subfigure}{0.255\linewidth}
        \centering
        \includegraphics[width=\linewidth]{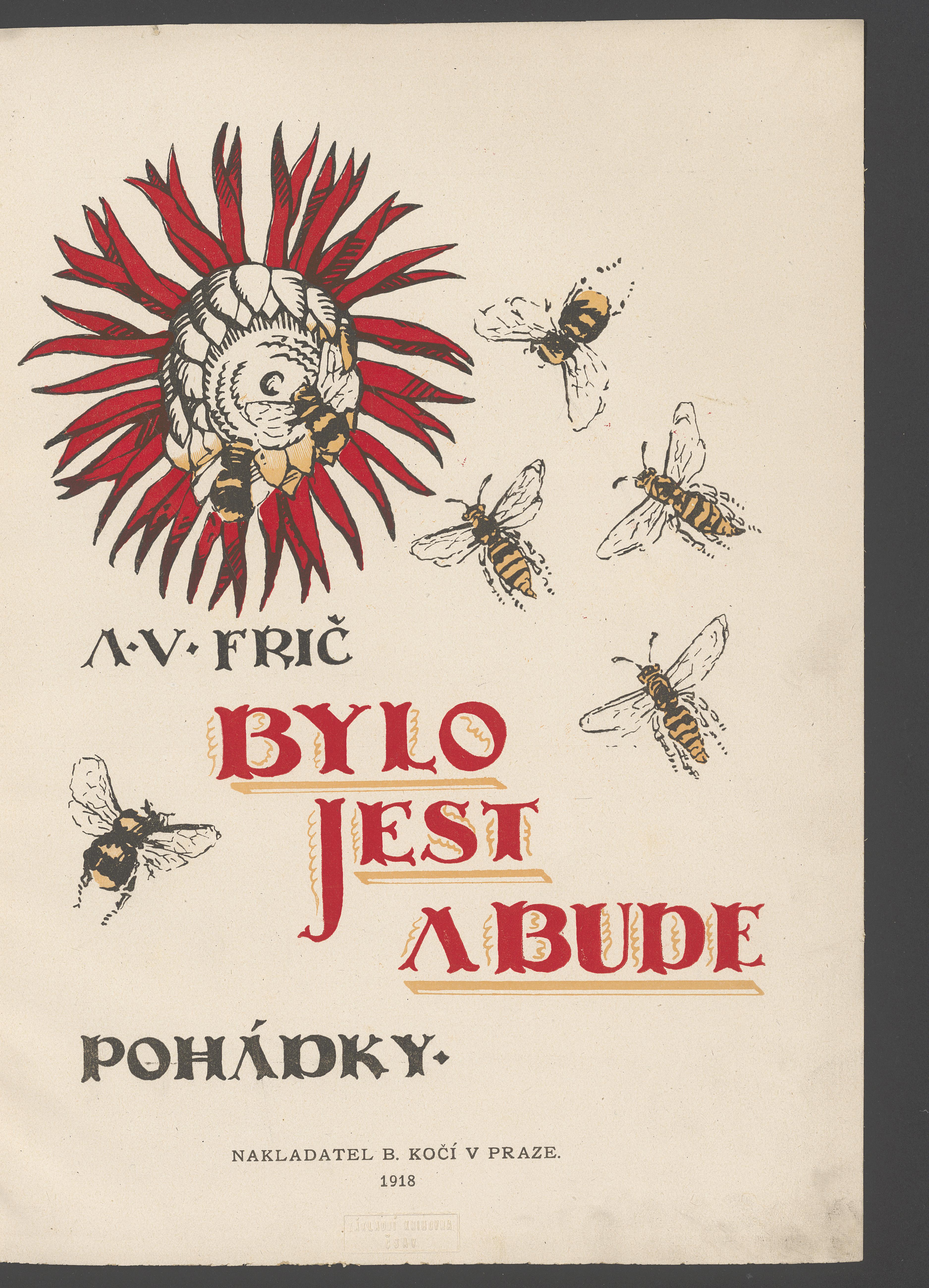}
    \end{subfigure}
    \begin{subfigure}{0.27\linewidth}
        \centering
        \includegraphics[width=\linewidth]{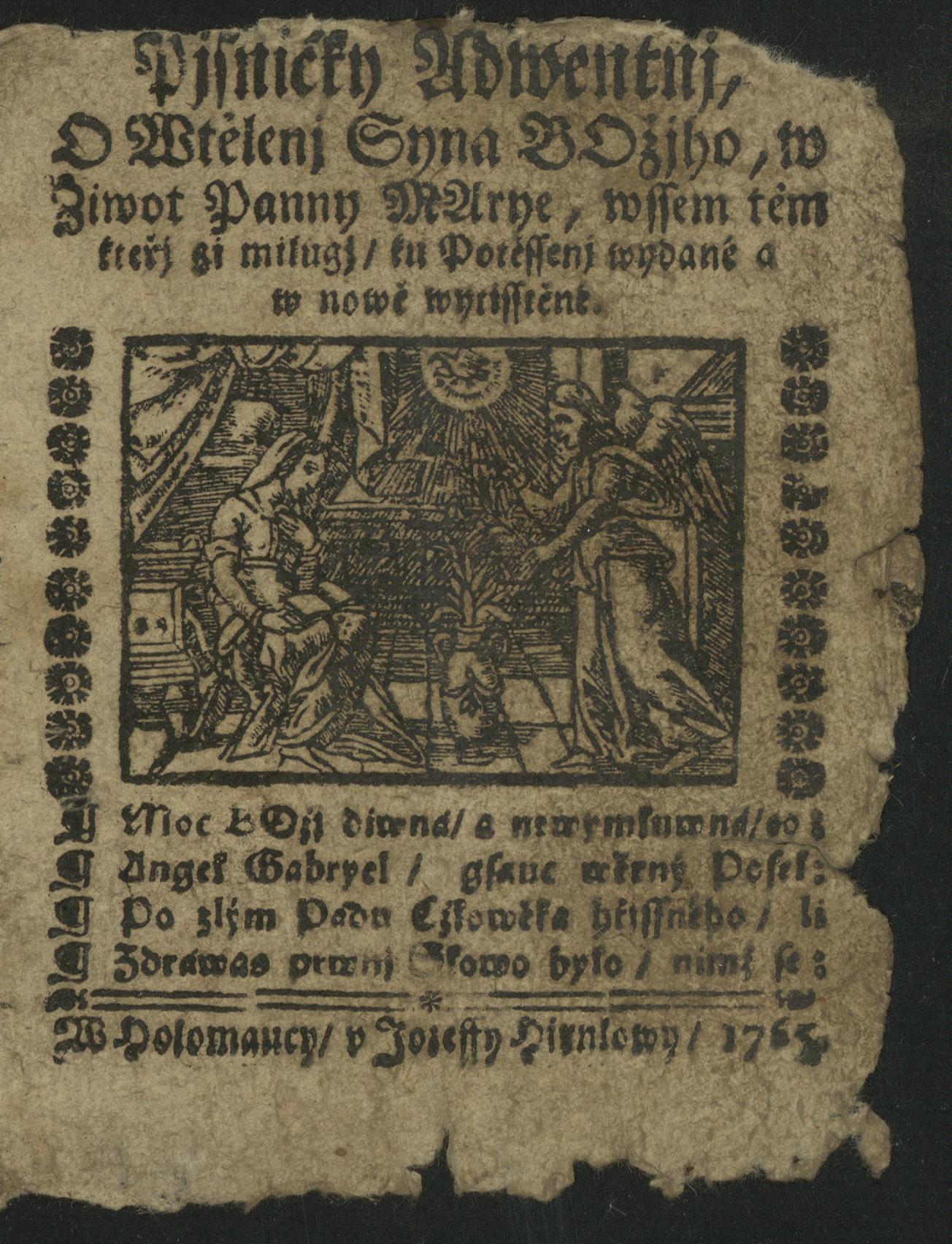}
    \end{subfigure}
    \caption{Representative title pages from BiblioPage dataset, selected to illustrate the various types found within it.}
    \label{fig:samples}
\end{figure}

The BiblioPage dataset comprises 2,118 annotated title pages, primarily from standalone works and multi-volume collections, sourced from 14 Czech libraries, with issue dates ranging from 1485 to the 21st century.
Across these pages, a total of 12,703 attributes were annotated, spanning 16 categories that capture various types of bibliographic information. 
The attributes include details related to the title, publication and edition, publisher and manufacturing, as well as contributor information.
Apart from the actual values of the attributes, we also include information about the geometrical position.

Figure~\ref{fig:samples} shows four representative title pages illustrating the scope and variability of our dataset.

The first page represents approximately 70\%~of the dataset, consisting of relatively simple title pages that present fundamental bibliographic details—such as the title, author, publisher, issue date, and place of publication—in a straightforward manner.
The second page accounts for about 25\%~of the dataset and contains more complex title pages with an increased amount of text, making information extraction more challenging. These pages often include additional bibliographic details, such as serial numbers or multi-set collection identifiers. In some cases, the expected structure is disrupted—for example, the name of a part may appear more prominently than the title, or various contributor roles may be mentioned indirectly.
The last two pages represent the remaining 5\%~of the dataset and consist of particularly challenging and visually distinctive examples. These pages either feature elaborate calligraphy and handcrafted designs or belong to historical prints with unique typographical styles.

\subsection{Data Sourcing and Distribution}

\begin{figure}[t]
    \centering
    \includegraphics[width=\linewidth, trim=0mm 4mm 0mm 0mm, clip]{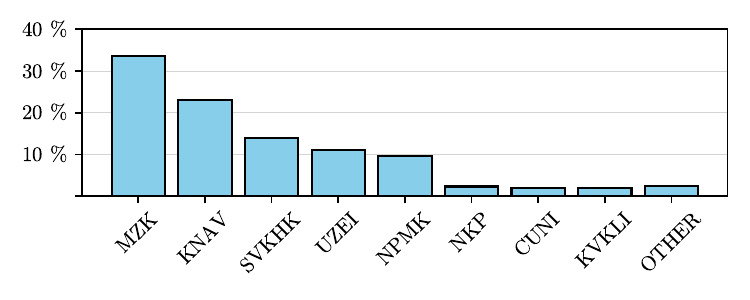}
    \caption{The distribution of BiblioPage dataset documents' over libraries.}
    \label{fig:dataset:libraries}
\end{figure}

\begin{figure}[t]
    \centering
    \includegraphics[width=\linewidth, trim=0mm 4mm 0mm 0mm, clip]{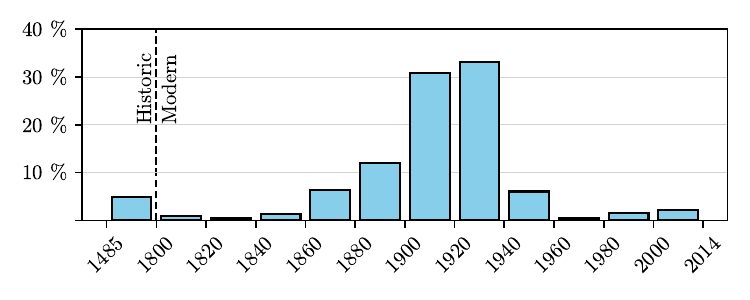}
    \caption{The distribution of BibliPage dataset documents' issue dates over time.}
    \label{fig:dataset:dates}
\end{figure}

We randomly sampled single-volume and multi-volume works from 14 Czech libraries. 
We excluded periodicals such as newspapers, magazines, and journals, as our focus is not on the bibliographic information specific to these types of documents. 
This selection process ensured a diverse and representative dataset, encompassing a broad spectrum of publication types across various disciplines and historical periods.
Figure~\ref{fig:dataset:libraries} illustrates the distribution of libraries, while Figure~\ref{fig:dataset:dates} depicts the distribution over time.

The Moravian Library (MZK) contributed the largest share of the dataset, which is consistent with its status as one of the largest libraries in the Czech Republic. 
Its collections include both fiction and non-fiction, as well as rare historical prints. 
Although the National Library of the Czech Republic (NKP) is also one of the country's largest libraries, its representation in our dataset is comparatively smaller. 
This is due to a significant overlap in collections between NKP and MZK, combined with our sampling preference for MZK in cases of duplication. 
The Library of the Czech Academy of Sciences (KNAV) and the Research Library in Hradec Králové (SVKHK) primarily focus on scientific resources, contributing a significant number of non-fiction publications. 
The Library of the Institute of Agricultural Economics and Information (UZEI) specialize in agriculture-related publications. 
The National Pedagogical Museum and Library of J. A. Comenius (NPMK) focuses on literature related to the development of Czech education within the European context and includes archival documents by classical pedagogical authors. 
The Charles University (CUNI) Library primarily contributed publications related to law.
The Regional Research Library in Liberec (KVKLI) has a distinct emphasis on regional history and specialized collections. 
It houses historical German-language publications from the Czech lands, as well as a historic collection featuring manuscripts, rare books, and archival materials.

A few other libraries contributed less than 20 pages to the dataset. These include the First Faculty of Medicine, Charles University (CUNI LF1), Museum Jindřichohradecka (MJH), National Film Archive (NFA), National Institute of Folk Culture (NULK), and The Academy of Arts, Architecture and Design in Prague (VŠUP).

Ninety percent of the works in our dataset were published within the 100-year span between 1860 and 1960, while the remaining 10\% includes both older historical prints and modern publications. 
This distribution reflects the relative scarcity of older works and the restricted public access to more recent publications.

\begin{figure}[t]
    \centering
    \begin{subfigure}{0.48\linewidth}
        \centering
        \includegraphics[width=\linewidth]{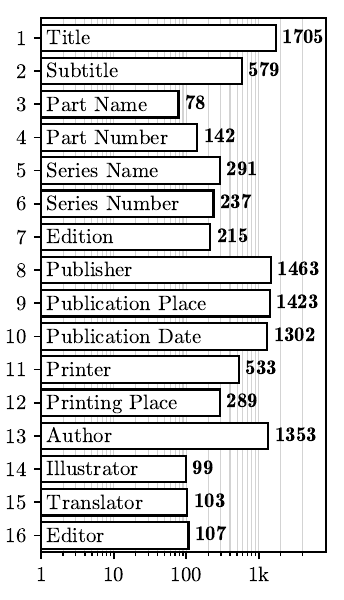}
        \caption{Development set distribution}
        \label{fig:distribution_train}
    \end{subfigure}
    \hfill
    \begin{subfigure}{0.48\linewidth}
        \centering
        \includegraphics[width=\linewidth]{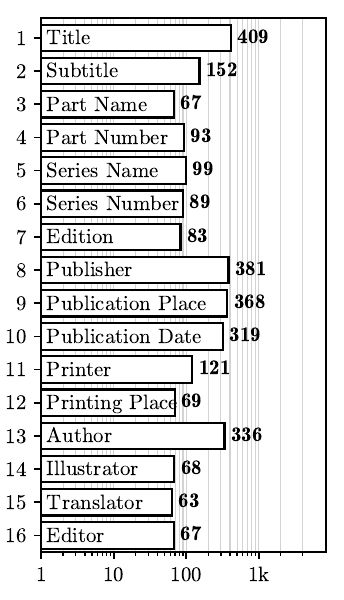}
        \caption{Test set distribution}
        \label{fig:distribution_test}
    \end{subfigure}
    \caption{Attribute distribution in the BiblioPage dataset for the development and test sets. A logarithmic scale is applied to the x-axis for visualization purposes.}
    \label{fig:attributes_distribution}
\end{figure}
When splitting the dataset into development and test subsets, the primary goal was to ensure that the test set adequately covered all 16 attributes. 
Since the attribute distribution is non-uniform and the number of pages is limited, simple uniform sampling of pages would result in insufficient representation for some attributes.
Therefore, we first randomly sampled 200 pages for the test set and then repeatedly sampled additional pages from the remaining ones for attributes that did not meet the predefined minimum coverage threshold of 50 instances.
This process resulted in 1,708 development pages and 410 test pages, with the final number of attributes totaling 9,919 and 2,784 respectively. Figure~\ref{fig:attributes_distribution} presents the distribution of the attributes. 

The Title and Subtitle identify the work. For works that are part of a larger collection, the Part Name and Part Number specify the section or volume, while the Series Name and Series Number indicate the work’s position within a series.
Publication details include the Edition, along with the Publication Place and Publication Date, which record where and when the work was released. 
Information about production and distribution includes the Publisher, Printer, and Printing Location.
Contributor details encompass four roles—Author, Illustrator, Translator, and Editor.

\subsection{Annotation Process}\label{subsec:annotation}
We adopted a semi-automatic approach to annotation, combining high quality OCR models (more details can be found in Section~\ref{sec:baseline}) for text recognition with bibliographic metadata obtained from the official catalogs of the respective libraries.
We localized the metadata in the OCR outputs to create annotation proposals which were then manually verified and corrected where needed. 
This process improved annotation efficiency, but, perhaps more importantly, having the official metadata as reference improved annotation quality in ambiguous cases.
The objective was to annotate all relevant bibliographic information on each title page, including its geometrical position. 
Specifically, for each bibliographic element in a page, the annotation includes the correct attribute type, a bounding box that indicates its exact location on the page, and the textual value of the attribute.

\subsubsection{Attribute annotation.} 
We annotated the precise position of each bibliographic element on the page along with its corresponding attribute.
To enhance reliability, we aligned OCR transcriptions with metadata from library catalogs, providing an initial estimate of each element’s location and attribute type.
However, this process had several limitations: library catalogs contained errors, OCR outputs were sometimes inaccurate, and the alignment itself occasionally failed to detect or correctly position certain elements. 
As a result, the alignment served as a guideline rather than a definitive source of positional accuracy. 
When it was unsuccessful but the bibliographic attribute was present in the catalog, we included a small annotation on the left side of the page to assist annotators. 
To further support accurate annotation, annotators were provided not only with suggested positions and attribute types but also with the textual content of the bibliographic information retrieved from the catalog.

We manually reviewed the alignment to ensure that the intended bibliographic information was accurately captured and corrected any misalignments. 
The first two images in Figure~\ref{fig:dataset:annotation} illustrate the automatic alignment of bibliographic attributes from the library catalog using OCR, followed by manual adjustments. Different colors represent different attributes, while unaligned suggested attributes appear to the left of the central illustration. For example, the pink annotation corresponding to the part number was present on the page but was not properly aligned, likely due to a discrepancy in format—while the catalog stored it as a numeric value ("2"), it appeared on the title page as a string ("second", "druhý" in Czech).

Using bounding boxes, we precisely annotated each bibliographic attribute, ensuring the relevant text was fully enclosed. 
However, due to the rectangular shape of the bounding boxes, precise annotation was sometimes not possible when text elements had irregular layouts. 
In such cases, the bounding box unavoidably included irrelevant text, which we retained while noting the affected pages for later correction during the final corrections phase.
Detailed rules for attribute annotations are given at the end of this section.

A team of five annotators carried out the task—four performed the annotations, while one reviewer systematically verified all results for consistency and accuracy.

\subsubsection{Text extraction.} 
\begin{figure}[t]
    \centering
    \begin{subfigure}{0.24\linewidth}
        \centering
        \includegraphics[width=\linewidth]{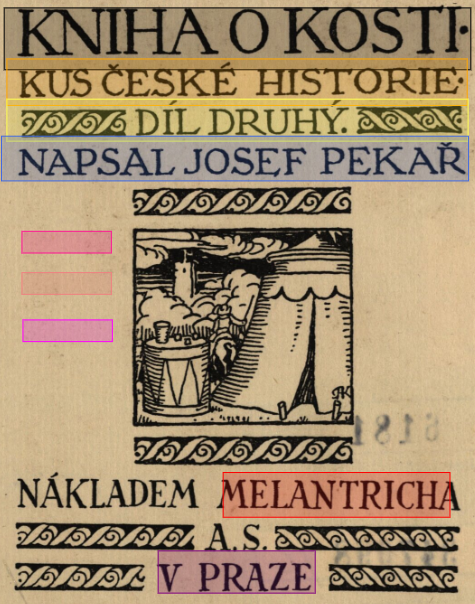}
    \end{subfigure}
    \hfill
    \begin{subfigure}{0.24\linewidth}
        \centering
        \includegraphics[width=\linewidth]{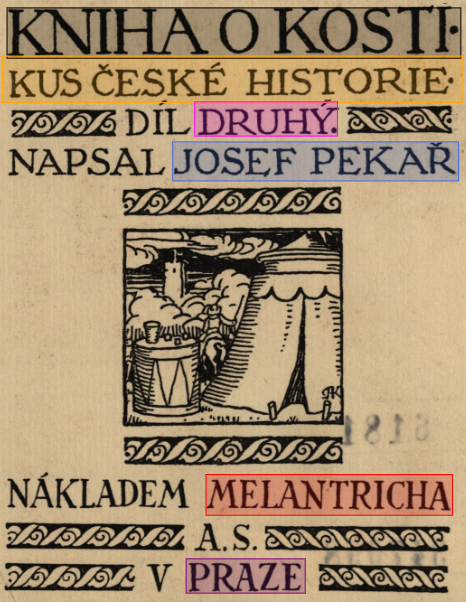}
    \end{subfigure}
    \hfill
    \begin{subfigure}{0.24\linewidth}
        \centering
        \includegraphics[width=\linewidth]{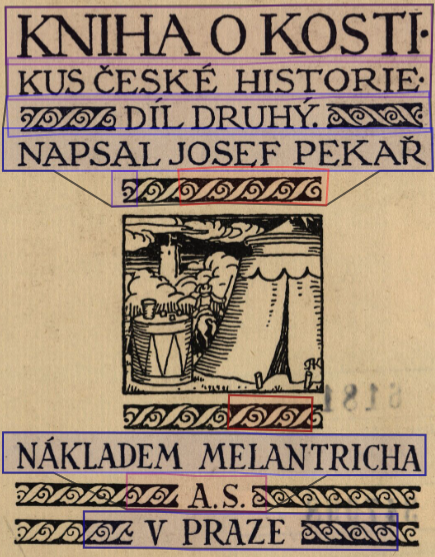}
    \end{subfigure}
     \hfill
    \begin{subfigure}{0.24\linewidth}
        \centering
        \includegraphics[width=\linewidth]{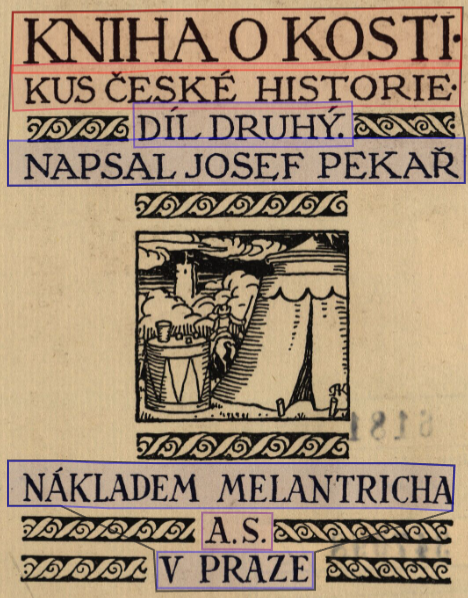}
    \end{subfigure}
    
    \caption{Automatic alignment of bibliographic information and its correction (left), automatic layout detection and its correction (right).}
    \label{fig:dataset:annotation}
\end{figure}

We extracted both text transcriptions and their geometric positions. 
We selected the PERO OCR~\cite{kodym2021page} for this task due to its text output quality and ability to transcribe historical typefaces.
Since accurate layout detection is crucial for reliable text extraction, we manually corrected this step to ensure all relevant text was captured, as illustrated in the last two images of Figure~\ref{fig:dataset:annotation}.
For most title pages, the PERO OCR system achieved high accuracy, even on historical pages using Fraktur typefaces.
However, the system encountered difficulties with highly non-standard fonts, particularly on calligraphic title pages with ornamental text. 
These cases required additional manual intervention to ensure proper transcription.

\subsubsection{Aligning attributes to the extracted text.}
After completing the attribute annotation and text extraction, we obtained two key positional sources: attribute positions from bounding boxes and text positions from OCR. 
We then aligned these two sources to generate the final annotations for review. 
Since both attribute bounding boxes and OCR layout detection were manually corrected, this alignment resulted in highly precise annotations, although some final checks were still necessary.

 
We manually reviewed the text annotations for all 410 title pages in the test subset of the dataset. Additionally, since we had identified problematic pages during the attribute annotation process, and the OCR system provided confidence scores for its transcriptions, we systematically corrected all low-confidence and problematic pages in the development set as well.

The final annotation for each page includes both textual and positional information related to the bibliographic content. The textual information is structured as a JSON dictionary, where each key represents an attribute name, and each value is either a single text entry or a list of texts, depending on the number of instances of the attribute on the page. Positional information is provided as bounding boxes in the YOLO format.

\subsubsection{Annotation rules for attributes.}
To simplify the annotation rule for attributes, we define it as the exact text printed on the title page. 
This means the attribute appears in the annotation in its original form, including its original declension. 
Numbers are always written as they appear, rather than converted to their numerical equivalent, and names may not always be in their nominative form.
While this approach clarifies what the annotation should contain, it may pose additional challenges for VLLM and LLM models processing our dataset.

For some attributes, additional guidelines were necessary. 
We explicitly excluded academic titles for contributors, publishers, and printers. 
If a publication place appeared multiple times referring to the same location, we annotated only the more visually dominant instance. 
This typically applied to cases where a city name appeared in capital letters at the bottom, conflicting with a smaller accompanying reference near the publisher’s name. 
For Part Number, Series Number, and Edition, we annotated only the numerical information, omitting any descriptive text following the number. 
The exception was when no numerical information was present, in which case we annotated the entire phrase, such as "Special Edition".

An important note for all attributes is that we did not annotate any general text corresponding to the attribute. 
For example, in the phrase "Special Edition," only the word "Special" is assigned to the attribute Edition, as the word "Edition" is irrelevant. 
This approach ensures that trained systems ideally produce bibliographic information without including general descriptive text related to the attribute itself.
This is because we want the trained systems to ideally produce the bibliographic information as such without including general descriptive text related to the attribute itself.

Multiple instances of the same attribute for a single page were annotated only for Publisher and contributor-realted attributes, as nearly all other attributes appeared only once per page. 
This distinction may be relevant for imposing prior constraints when developing models for this dataset.

\subsection{Evaluation protocol}
As described in the previous section, the BiblioPage dataset is structured as a JSON dictionary for each page, where each key corresponds to one of the 16 attributes. The values can be either a single entry or multiple elements, depending on the number of occurrences of the attribute on the page.

For each page and attribute, the evaluation of proposed values follows a matching procedure. Each ground truth value is compared against each proposed value, ensuring that one ground truth value can match at most one proposed value. Matched values are counted as true positives, unmatched proposed values as false positives, and unmatched ground truth values as false negatives. 
A match is considered valid if the Character Error Rate (CER) does not exceed 10\% when compared to the ground truth.
Proposed values may include confidence scores, allowing for the computation of mean Average Precision (mAP) and other metrics at different confidence thresholds. To facilitate evaluation, we provide a script that processes two folders containing JSON files, one for proposed values and one for ground truth, automatically performing the comparison and metric computation.

To eliminate penalization of irrelevant characters, we apply normalization to all bibliographic metadata before evaluating. 
We remove super rare characters, trim whitespace, and collaps multiple whitespaces into one. 
Additionally, historical characters are replaced with modern equivalents, such as converting long s to s, æ to ae, or replacing old german umlaut representations with their modern equivalents. 
Additionally, all attributes except Title, Subtitle, Part Name, and Series Name undergo further normalization by stripping punctuation and quotes for consistency.

\section{Baselines}\label{sec:baseline}

We evaluated two types of bibliographic attribute extraction approaches on our dataset.

The first type combines object detection with OCR. 
The object detector predicts the position and class of each attribute, while the OCR extracts the corresponding text. 
We evaluate various object detection models within this framework, using the detector’s confidence score as the final prediction confidence, allowing for flexible configurations in standard object detection pipelines.

The second approach leverages Vision-Language Large Models (VLLMs) in a zero-shot setting, where only instructions, images and optionally OCR outputs are provided. 
These instructions are highly detailed, specifying the dataset's annotation structure and including representative text examples for each attribute from the development subset. 
While this approach does not follow the few-shot paradigm, where models are explicitly provided with multiple examples during inference, it still benefits from structured guidance about the dataset.

\subsection{Object detection}

We used the YOLO Ultralytics package~\cite{yolo11_ultralytics} to train the nano, small, medium, and large versions of YOLOv11.
We disabled translation and flipping augmentations. 
Translation augmentation can shift annotated bounding boxes partially outside the image, resulting in incomplete text instances (e.g., only a few letters of a name). 
Since the model is expected to differentiate full names and other bibliographic attributes, such partial annotations could introduce ambiguity and negatively impact learning. 
Flipping is disabled because the direction of the text is meaningful.
To estimate the optimal number of training epochs, we conducted a preliminary experiment for each model size on the development subset, using 200 pages for validation and the rest for training. 
Based on the best-performing checkpoint, we set the epoch count for the final training on the full development set, increasing it by 10\% to account for increased size of the training set.

We incorporated two OCR models: one for alignment and another for transcription. 
The alignment model is based on connectionist temporal classification (CTC) with a convolutional and recurrent architecture, the alignment utilizes the per-frame output of CTC-based models. 
The transcription model is transformer-based with a CNN backbone, trained on a large-scale dataset spanning both printed and handwritten material from historical and modern periods, including IMPACT~\cite{IMPACT2013} and the Deutsches Textarchiv (DTA)\footnote{https://www.deutschestextarchiv.de}. This ensures high-accuracy transcription of BiblioPage title pages, including older Fraktur prints.

The final outputs were obtained by aligning YOLO and OCR predictions using the approach described in Section~\ref{subsec:annotation}. If an object detection does not correspond to any recognized text, it is discarded. Alternatively, an extension to our approach could enhance performance by explicitly reprocessing these misaligned detections with OCR, enabling the system to capture text that may have been missed due to layout analysis errors.

\begin{figure}[t]
    \centering
    \includegraphics[width=\linewidth, trim=0mm 4mm 0mm 0mm, clip]{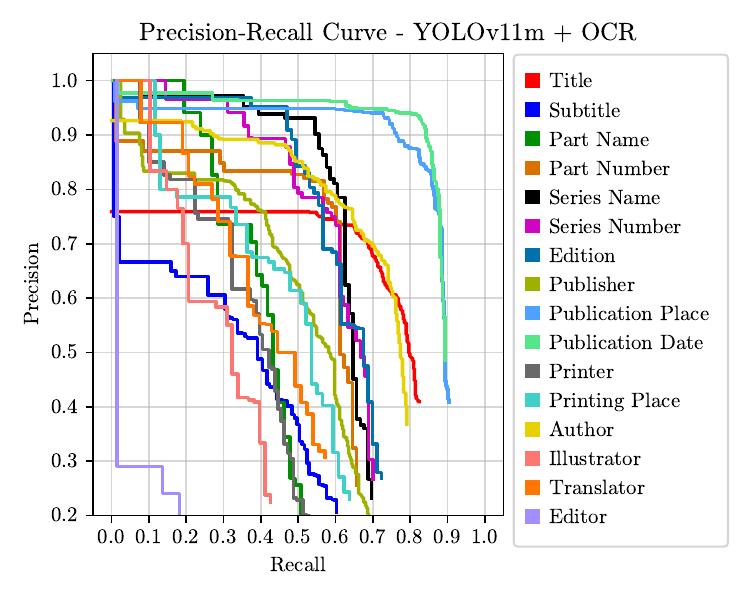}
    \caption{Precision-Recall curves across 16 attributes found in BiblioPage dataset.}
    \label{fig:baseline:yolov11m}
\end{figure}

\begin{table}[t]
\caption{Performance for various sizes of YOLO models combined with OCR.}\label{tab:baseline:precision_recall_detail}
\centering
{
\begin{tabular}{ r | c | g | g | c| c| c| c| c| c| g| g| c| c| g| c| c | c | l
}
 & \rotatebox[origin=l]{90}{Mean} &\rotatebox[origin=l]{90}{Title} & \rotatebox[origin=l]{90}{Subtitle} & \rotatebox[origin=l]{90}{Part Name} & \rotatebox[origin=l]{90}{Part Number} & \rotatebox[origin=l]{90}{Series Name} & \rotatebox[origin=l]{90}{Series Number} & \rotatebox[origin=l]{90}{Edition} & \rotatebox[origin=l]{90}{Publisher} & \rotatebox[origin=l]{90}{Publication Place} & \rotatebox[origin=l]{90}{Publication Date} & \rotatebox[origin=l]{90}{Printer} & \rotatebox[origin=l]{90}{Printing Place} & \rotatebox[origin=l]{90}{Author} & \rotatebox[origin=l]{90}{Illustator} & \rotatebox[origin=l]{90}{Translator} & \rotatebox[origin=l]{90}{Editor}\\
\hline\hline
 F1 & 43 & 68 & 41 & 25 & 44 & 50 & 50 & 46 & 44 & 74 & 83 & 18 & 27 & 61 & 30 & 32 & 0 & YOLOv11n \\
 P & 50 & 63 & 38 & 55 & 57 & 51 & 56 & 50 & 41 & 69 & 87 & 46 & 42 & 54 & 56 & 31 & 0\\
 R & 41 & 73 & 43 & 16 & 36 & 49 & 45 & 43 & 47 & 79 & 78 & 11 & 20 & 69 & 20 & 33 & 0 \\
 AP & 38 & 62 & 34 & 16 & 37 & 44 & 43 & 42 & 38 & 80 & 80 & 12 & 18 & 64 & 18 & 21 & 0 \\
\hline
F1 & 49 & \textbf{70} & 44 & 43 & 52 & 59 & 51 & 54 & 49 & 80 & 84 & 22 & 36 & 66 & 20 & 43 & 10 & YOLOv11s\\
P & 56 & 67 & 42 & 53 & 73 & 57 & 67 & 66 & 48 & 78 & 87 & 33 & 50 & 63 & 47 & 56 & 19 \\
R & 45 & 72 & 46 & 37 & 40 & 60 & 41 & 46 & 50 & 81 & 81 & 16 & 28 & 70 & 13 & 34 & 7 \\
AP & 44 & 61 & 39 & 34 & 46 & 55 & 48 & 49 & 45 & 82 & 83 & 11 & 30 & 66 & 21 & 31 & 3 \\
\hline
F1 & \textbf{57} & 67 & 43 & \textbf{46} & \textbf{66} & \textbf{65} & 62 & 64 & 54 & \textbf{84} & 86 & \textbf{46} & 52 & 69 & \textbf{40} & 45 & \textbf{18} & YOLOv11m \\
P & 63 & 63 & 46 & 50 & 81 & 69 & 79 & 80 & 54 & 83 & 87 & 55 & 64 & 64 & 58 & 54 & 29 \\
R & 53 & 72 & 40 & 43 & 55 & 62 & 51 & 54 & 54 & 84 & 85 & 39 & 44 & 74 & 30 & 39 & 13 \\
AP & 50 & 59 & 31 & 39 & 52 & 61 & 59 & 60 & 48 & 83 & 84 & 37 & 44 & 66 & 27 & 39 & 4 \\
\hline
F1 & \textbf{57} & \textbf{70} & \textbf{50} & 45 & 61 & \textbf{65} & \textbf{64} & \textbf{69} & \textbf{57} & 82 & \textbf{87} & 40 & \textbf{55} & \textbf{70} & 34 & \textbf{47} & 10 & YOLOv11l \\
P & 66 & 67 & 57 & 58 & 76 & 71 & 82 & 79 & 57 & 81 & 90 & 44 & 76 & 67 & 64 & 75 & 14 \\
R & 51 & 73 & 45 & 37 & 51 & 60 & 53 & 61 & 56 & 82 & 84 & 37 & 43 & 73 & 23 & 34 & 7 \\
AP & 50 & 60 & 39 & 35 & 53 & 59 & 58 & 65 & 51 & 81 & 85 & 33 & 48 & 63 & 29 & 34 & 1 \\

\end{tabular}}
\end{table}

Figure~\ref{fig:baseline:yolov11m} shows the precision-recall curves for each attribute, where the object detector’s confidence score is used as the system’s confidence.
The detector employed is the medium version of YOLOv11.
Table~\ref{tab:baseline:precision_recall_detail} presents precision (P), recall (R), F1 score (F1), and average precision (AP) for each attribute, along with the mean across all attributes. 
The confidence threshold for the reported values was determined as the point where the averaged F1 curve reached its maximum value.

There was no significant difference in performance for the attributes Title, Subtitle, Publication Place, Publication Date, and Author (highlighted with a gray background) across different model sizes. 
Among these, Publication Place and Publication Date were the easiest to predict accurately, likely because they were consistently positioned near the bottom of title pages while at the same time being visually distinct.

The performance on Title, Subtitle, and Author, while not perfect, remained stable across different model sizes. 
This suggests a mix of highly distinguishable and challenging instances.
Some title pages feature clear title and author information, while others contain dense text, making it difficult to identify the correct attributes.
Additionally, when multiple contributors are listed, distinguishing the Author role can be challenging.

The most difficult attributes to predict were contributor-related roles other than Author, likely due to their underrepresentation in the dataset. 
Recognizing Subtitle was also challenging, as its distinction from Title is not always clear, and not all text appearing below the title qualifies as a subtitle. 
Similarly, Part Name recognition heavily depends on the semantic meaning of the text within the title page, making it difficult to classify based on visual characteristics alone.

Overall, the YOLO models combined with OCR provide a reasonable baseline. 
Their primary limitation is their inability to leverage the semantic meaning of the text, as OCR operates separately from the classification process. 
Additionally, the number of annotated title pages may be insufficient for the detector to fully learn the semantic relationships between page entities necessary for more accurate predictions.

\subsubsection{DETR.} In light of recent advances in end-to-end detection transformers (DETR)~\cite{vedaldi_end--end_2020}, we conduct an ablation study on several representative DETR variants. 
The setup is the same as for YOLO-based systems, only now various DETR-based models are used for object detection.

Transformer-based detectors eliminate the need for anchors and non-maximum suppression (NMS) by directly predicting bounding boxes. Deformable DETR~\cite{zhu_deformable_2020} introduced multi-scale deformable attention, which improves small-object detection and achieves roughly $10\times$ faster training convergence than the original DETR. Subsequent works like Conditional DETR~\cite{meng_conditional_2021} and DAB-DETR~\cite{liu_dab-detr_2021} further refined the query and cross-attention mechanisms to accelerate training convergence significantly. More recently, RT-DETR~\cite{zhao_detrs_2024} incorporated an efficient hybrid image encoder to prioritize real-time detection. 

We evaluated all these DETR-based models integrated with our OCR pipeline and present the results in Table~\ref{tab:baseline:precision_recall}.
The results show significantly lower performance compared to YOLO-based systems, likely due to the limited size of our dataset. This suggests that DETR-based detectors are less suitable for small-scale datasets.

\subsection{VLLM}

\begin{table}[t]
\caption{Per attribute performance for selected VLLM models. Columns are sorted by average F1 in descending order.}\label{tab:baseline:precision_recall_vllm}
\centering
{
\begin{tabular}{r | c | c| c| c| c| c| c| c| c| c| c| c| c| c| c| c | c | l
}
& \rotatebox[origin=l]{90}{Average} & \rotatebox[origin=l]{90}{Publication Date} & \rotatebox[origin=l]{90}{Illustator} & \rotatebox[origin=l]{90}{Publication Place} & \rotatebox[origin=l]{90}{Translator} & \rotatebox[origin=l]{90}{Author} & \rotatebox[origin=l]{90}{Title} & \rotatebox[origin=l]{90}{Series Name} & \rotatebox[origin=l]{90}{Part Number} & \rotatebox[origin=l]{90}{Printing Place} & \rotatebox[origin=l]{90}{Series Number} & \rotatebox[origin=l]{90}{Edition} & \rotatebox[origin=l]{90}{Subtitle} & \rotatebox[origin=l]{90}{Publisher} & \rotatebox[origin=l]{90}{Editor} & \rotatebox[origin=l]{90}{Part Name} & \rotatebox[origin=l]{90}{Printer} \\
\hline\hline
F1     & 67                                                     & \textbf{96}                                                       & 83                                                          & 89                                                                 & 92                                                          & 81                                                      & \textbf{62}                                            & \textbf{77}                                                  & \textbf{73}                                                  & 66                                                              & \textbf{71}                                                    & 64                                                       & \textbf{47}                                               & 44                                                         & \textbf{57}                                             & 37                                                         & 38                                                       & GPT-4o                     \\
P      & 69                                                       & 95                                                                & 84                                                          & 89                                                                 & 95                                                          & 80                                                      & 62                                                     & 79                                                           & 68                                                           & 73                                                              & 92                                                             & 64                                                       & 36                                                        & 43                                                         & 61                                                      & 40                                                         & 41                                                       &                            \\
R      & 67                                                       & 97                                                                & 82                                                          & 89                                                                 & 90                                                          & 83                                                      & 62                                                     & 76                                                           & 80                                                           & 60                                                              & 57                                                             & 65                                                       & 67                                                        & 45                                                         & 54                                                      & 35                                                         & 35                                                       &                            \\ \hline
F1     & \textbf{70}                                              & \textbf{96}                                                       & \textbf{94}                                                 & \textbf{91}                                                        & \textbf{94}                                                 & \textbf{84}                                             & 60                                                     & \textbf{77}                                                  & 72                                                           & \textbf{79}                                                     & 65                                                             & \textbf{68}                                              & 44                                                        & \textbf{56}                                                & 51                                                      & \textbf{49}                                                & \textbf{49}                                              & GPT-4o + OCR               \\
P      & 72                                                       & 95                                                                & 95                                                          & 93                                                                 & 96                                                          & 82                                                      & 60                                                     & 80                                                           & 64                                                           & 78                                                              & 84                                                             & 67                                                       & 35                                                        & 55                                                         & 54                                                      & 56                                                         & 52                                                       &                            \\
R      & 71                                                       & 97                                                                & 94                                                          & 90                                                                 & 92                                                          & 86                                                      & 60                                                     & 74                                                           & 81                                                           & 79                                                              & 53                                                             & 69                                                       & 60                                                        & 57                                                         & 48                                                      & 43                                                         & 47                                                       &                            \\ \hline
F1     & 8                                                        & 34                                                                & 7                                                           & 9                                                                  & 3                                                           & 15                                                      & 26                                                     & 6                                                            & 5                                                            & 3                                                               & 7                                                              & 0                                                        & 8                                                         & 3                                                          & 0                                                       & 6                                                          & 0                                                        & Llama 3.2-Vision      \\
P      & 9                                                        & 39                                                                & 6                                                           & 12                                                                 & 4                                                           & 14                                                      & 27                                                     & 11                                                           & 5                                                            & 3                                                               & 10                                                             & 0                                                        & 7                                                         & 3                                                          & 0                                                       & 5                                                          & 0                                                        &                           11B  \\
R      & 8                                                        & 30                                                                & 8                                                           & 8                                                                  & 3                                                           & 17                                                      & 26                                                     & 5                                                            & 5                                                            & 2                                                               & 5                                                              & 0                                                        & 9                                                         & 3                                                          & 1                                                       & 7                                                          & 0                                                        &                            \\ \hline
F1     & 29                                                       & 72                                                                & 63                                                          & 52                                                                 & 43                                                          & 47                                                      & 46                                                     & 13                                                           & 15                                                           & 14                                                              & 6                                                              & 3                                                        & 24                                                        & 21                                                         & 17                                                      & 13                                                         & 10                                                       & Llama 3.2-Vision\\
P      & 32                                                       & 72                                                                & 67                                                          & 53                                                                 & 50                                                          & 45                                                      & 48                                                     & 23                                                           & 11                                                           & 46                                                              & 12                                                             & 2                                                        & 18                                                        & 18                                                         & 25                                                      & 11                                                         & 17                                                       &                            11B + OCR \\
R      & 29                                                       & 72                                                                & 60                                                          & 51                                                                 & 38                                                          & 50                                                      & 43                                                     & 9                                                            & 25                                                           & 8                                                               & 4                                                              & 4                                                        & 34                                                        & 24                                                         & 13                                                      & 16                                                         & 7                                                        &                            \\ \hline\hline
AVG F1 & -                                                        & 75                                                                & 62                                                          & 60                                                                 & 58                                                          & 57                                                      & 49                                                     & 43                                                           & 41                                                           & 41                                                              & 37                                                             & 34                                                       & 31                                                        & 31                                                         & 31                                                      & 26                                                         & 24                                                       &                           

\end{tabular}}
\end{table}

We conducted zero-shot experiments with various Visual Large Language Models (VLLMs). Specifically, we evaluated open-weight Llama 3.2-Vision models with 11B and 90B parameters~\cite{Dubey2024TheL3}, as well as the closed-source GPT-4o mini and GPT-4o models from OpenAI~\cite{Achiam2023GPT4TR,openai_api}.
Results are summarized in Tables~\ref{tab:baseline:precision_recall_vllm} and \ref{tab:baseline:precision_recall}. Inference for Llama models was performed using the Ollama server~\cite{ollama}.

We performed prompt tuning using a subset of the development data with the Llama 3.2-Vision 11B model, applying a multi-turn chat format.  Each prompt included attribute descriptions, hints of probable location within a page and Czech examples. 
Models were explicitly instructed to exclude general terms (e.g., "author," "illustrator"), academic titles, and unnecessary prepositions. 
However, they still often retained them. 
The models generated responses in JSON format.

We further examined the impact of including OCR output in prompts, explicitly mentioning its potential unreliability. Despite possible OCR errors, incorporating OCR significantly improved performance, particularly for the Llama models, mitigating their language support limitations (primarily Czech).

Our results indicate that attributes introduced with identifiers, such as Illustrator or Translator, or those like Publication Date, have a clearer form and perform better on average. Surprisingly, all models struggled with correctly identifying titles and subtitles, which are visually the most significant elements. Manual analysis showed errors commonly occurred when longer titles spanned multiple lines or featured varying font sizes. 
OCR input did not resolve this issue since it retains the original layout separation into individual text lines.

\begin{table}[t]
\caption{Comparison of all proposed baselines.}\label{tab:baseline:precision_recall}
\centering
{
\begin{tabular}{r | c | c | c | c }
 & mF1 & mP & mR & mAP \\
\hline\hline
YOLOv11n & 44 & 49 & 43 & 40\\
YOLOv11s & 52 & 59 & 49 & 47\\
YOLOv11m & 59 & 68 & 53 & \textbf{52}\\
YOLOv11l & 58 & 66 & 54 & 51\\
\hline
DABDETR & 24 & 22 & 35 & 24 \\
DeformableDETR & 16 & 19 & 22 & 15\\
DETR & 13 & 13 & 17 & 11\\
CoDETR & 11 & 10 & 18 & 11\\
RTDETR & 1 & 3 & 1 & 0 \\
\hline
Llama 3.2-Vision 11B & 8 & 9 & 8 & - \\
Llama 3.2-Vision 90B & 10 & 9 & 11 & - \\
GPT-4o mini & 41 & 43 & 41 & - \\
GPT-4o & 67 & 69 & 67 & - \\
\hline
Llama 3.2-Vision 11B (OCR) & 29 & 32 & 29 & - \\
Llama 3.2-Vision 90B (OCR) & 44 & 47 & 44 & - \\
GPT-4o mini (OCR) & 53 & 54 & 55 & - \\
GPT-4o (OCR) & \textbf{70} & \textbf{72} & \textbf{71} & - \\
\end{tabular}}
\end{table}

Comparing the VLLM-based approaches to the object detection-based ones, VLLMs clearly benefit from the semantic understanding of the text. This advantage makes attributes such as Translator and Illustrator easier to recognize, as their roles are often directly stated on title pages. 
Additionally, VLLMs achieve reasonable performance for the Editor attribute, despite its specification being almost exclusively indirect and highly variable.

Table~\ref{tab:baseline:precision_recall} offers comparison of all evaluated baseline approaches, including those not detailed in detail in Table~\ref{tab:baseline:precision_recall_detail} and Table~\ref{tab:baseline:precision_recall_vllm}.

\section{Conslusion}

We introduced BiblioPage, a dataset of scanned title pages annotated with structured bibliographic metadata, aiming to support the automatic extraction of bibliographic information. 
The dataset covers a diverse range of publications across different periods, typographic styles, and archival sources.

To evaluate the dataset, we implemented two baseline approaches: object detection combined with OCR and Vision-Language Large Models (VLLMs). 
Object detection methods, which rely purely on visual cues, perform well for attributes with clear typographic patterns but struggle with those requiring textual semantic interpretation, such as contributor roles. VLLMs, on the other hand, leverage textual meaning, making them effective in recognizing attributes like Translator, Illustrator, and Editor where roles are explicitly or implicitly stated. 
However, both approaches struggle with Title and Subtitle recognition due to their multi-line structure, varying font sizes, and inconsistent typesets. 
The lack of stable textual or visual indicators, together with their potential for mutual confusion, further complicates accurate identification.

YOLOv11 medium achieved an mF1 of 59, while GPT-4o outperformed it with mF1 of 67, further improving to mF1 of 70 when supplemented with OCR. DETR-based object detection approaches performed significantly worse than YOLO-based solutions, likely due to the limited size of our dataset. Similarly, LLaMA-based VLLMs underperformed compared to GPT-4o, possibly due to their weaker suitability for document extraction and OCR-related tasks, as well as their lower proficiency in Czech, the primary language of our dataset.

BiblioPage dataset contributes to document information extraction research by providing a real-world dataset that serves as a benchmark for related fields, enabling the evaluation of various extraction methods across various bibliographic entities. At the same time, it supports the development of automated bibliographic metadata processing for archives and libraries.

\begin{credits}
\subsubsection{\ackname} This work has been supported by the Ministry of Culture Czech
Republic in NAKI III project Machine learning for printed heritage digitisation
(DH23P03OVV066), and Ministry of Education, Youth and Sports of the Czech Republic through the e-INFRA CZ (ID:90254).
\end{credits}
%
%

\bibliographystyle{splncs04}
\bibliography{mybibliography}

\end{document}